\documentclass{article}
\usepackage{array}
\usepackage{colortbl}

 \usepackage[preprint]{corl_2026} 
\usepackage{amsmath,amssymb}
\usepackage{algorithm}
\usepackage{algpseudocode}
\usepackage{booktabs}
\usepackage[table]{xcolor}
\usepackage{graphicx}
\usepackage[normalem]{ulem}
\definecolor{best}{HTML}{FFC0CB} 

\usepackage{graphicx}
\usepackage{diagbox}
\usepackage{booktabs}
\usepackage{multirow}
\usepackage{tabularx}
\usepackage{subcaption}
\usepackage{tikz}

\usepackage[table]{xcolor}
\usepackage{graphicx}
\usepackage[normalem]{ulem}

\usepackage{enumitem}

\usepackage{wrapfig}

\title{DexFuture: Hierarchical Future-State Visuomotor Targeting for Bimanual Dexterous Tool Use}

\author{
\textbf{Runfa Blark Li, Kuang-Ting Tu, Nikola Raicevic, Dwait Bhatt, Xinshuang Liu, Keito Suzuki,}\\
\textbf{Ki Myung Brian Lee, Nikolay Atanasov, Truong Nguyen}\\
UC San Diego\\
}

\begin{document}
\maketitle

\vspace{-1cm}
\begin{figure}[h]
\centering
  \includegraphics[width=\textwidth]{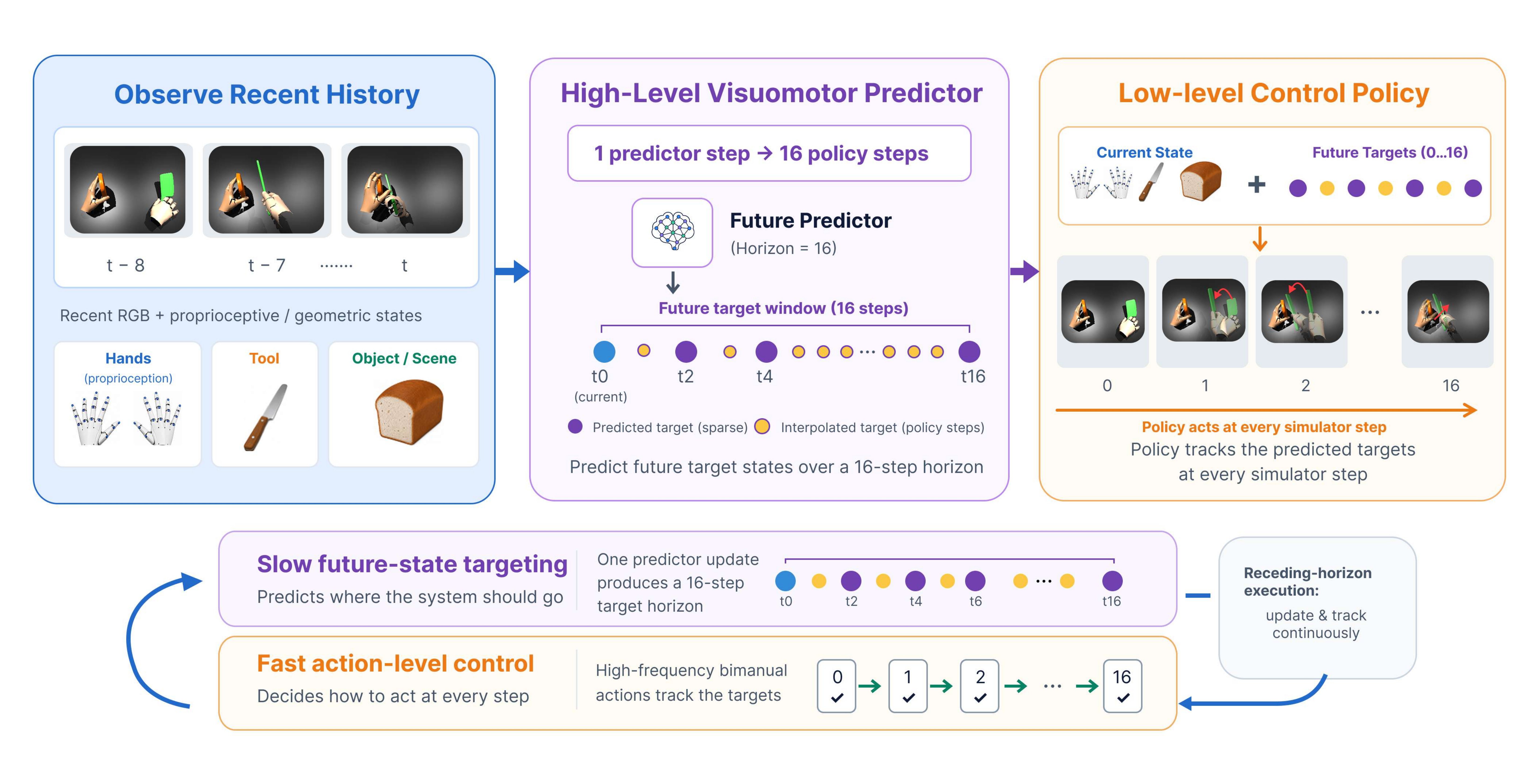}
  \vspace{-1cm}
  \captionof{figure}{\textbf{DexFuture} is a hierarchical system for bimanual dexterous tool use that couples a high-level \emph{Future-State Visuomotor Target Predictor} with a low-level \emph{Target-Conditioned Structured Dexterous Policy}. It removes the need for privileged future demonstration targets by predicting a coarse future-state target trajectory from visuomotor history. The predicted targets provide long-horizon guidance for the low-level policy to execute high-frequency contact-rich actions.}
  \label{fig:fig1}
\end{figure}

\vspace{-0.2cm}



\begin{abstract}
Bimanual dexterous tool use remains challenging for robots due to high-dimensional hand configurations and complex hand-tool-object dynamics and contact. 
Most existing control policies depend on future configuration references provided from demonstrations, while future action-conditioned world models require slow online planning over high-dimensional action sequences.
A significant challenge is generating a dynamically consistent future reference trajectory without relying on privileged states from demonstrations or slow counterfactual planning.
We propose \textbf{DexFuture}, a hierarchical system that couples a high-level \emph{Future-State Visuomotor Target Predictor} with a low-level \emph{Target-Conditioned Structured Dexterous Policy}. Conditioned on egocentric RGB, proprioceptive and geometric history, the high-level predictor constructs structured hand-tool-object visuomotor embeddings and uses a horizon-conditioned transformer to generate a multi-step future target trajectory. 
Then, the low-level policy tracks them with a target-conditioned per-link transformer. 
This hierarchy decouples coarse future reference generation from fine-grained action control, and slow long-horizon semantic prediction from high-frequency execution. On OakInk2 bimanual tool-use tasks, DexFuture achieves $90\%$ of the privileged-oracle performance, compared to $7\%$ for a no-reference policy. DexFuture operates at $60$ Hz, approximately $250\times$ faster than DexWM-style Cross-Entropy Method (CEM)
planning with a future action-conditioned world model. Project Website: \url{https://blarklee.github.io/DexFuture_official_website/}
\end{abstract}

\section{Introduction}
Bimanual dexterous tool use remains a central challenge in robot learning, requiring two high-DoF hands to coordinate indirect contacts through a tool while interacting with an object. Recent learning-based methods have made impressive progress by leveraging demonstrations as strong guidance for dexterous control. In particular, target-conditioned policies make high-DoF manipulation more tractable by providing the policy with future reference targets extracted from demonstration trajectories ~\cite{dexmachina,dextrack,omnigrasp,maniptrans,physgraph}. These targets can encode future hand motion, object pose, fingertip relations, and task-level cues. However, they are also privileged. At deployment, the robot observes the current scene and proprioception, but does not have access to the future demonstration state. Thus, a key bottleneck is generating a dynamically meaningful future target trajectory without relying on privileged future demonstrations.

A natural alternative is to learn an action-conditioned world model and use online planning to select future actions~\cite{dexwm}. This enables counterfactual rollouts, but requires optimizing over many high-dimensional candidate action sequences at inference. For dexterous tool use, such planning remains computationally prohibitive for the high frequency needed for stable contact-rich control. A plausible compromise is to retain a fast target-conditioned policy and make only the high-level target predictor action-conditioned. However, this exposes a dependency loop: future actions depend on predicted future targets, while the predictor itself would require future actions as input.
This motivates a different formulation: rather than planning over future actions online, can we predict the future target interface directly from visuomotor history while still preserving the control advantages of future targets?

We propose \textbf{DexFuture}, a hierarchical future-state visuomotor control framework for bimanual dexterous tool use, illustrated in Fig.~\ref{fig:fig1}. DexFuture couples a high-level Future-State Visuomotor Target Predictor with a low-level Target-Conditioned Structured Dexterous Policy. The predictor observes recent egocentric RGB frames and proprioceptive/geometric states, and predicts a coarse future target trajectory over multiple policy steps. The policy then tracks the predicted and interpolated targets at every step to produce high-frequency bimanual actions. This hierarchy separates \emph{what/where} the manipulation should progress toward in the future from \emph{how} to execute contact-rich actions, while also separating slow long-horizon prediction from fast per-step control.

We evaluate DexFuture on challenging bimanual tool-use tasks from OakInk2~\cite{oakink2}, including cutting, pouring, wiping, shearing, and stirring. DexFuture removes the need for privileged future demonstration targets while recovering most of the performance of the oracle target-conditioned policy. Compared with a no-target policy, it substantially improves task success, showing that future targets remain essential for dexterous tool use. Compared with DexWM-style CEM planning \cite{dexwm}, DexFuture executes at the policy control rate, highlighting the practical advantage of amortized target prediction over online action-sequence optimization.

Our contributions are summarized as follows:
\begin{itemize}[leftmargin=1.2em, itemsep=0em, topsep=0em, parsep=0em, partopsep=0em]
\item We propose \textbf{DexFuture}, a hierarchical future-state visuomotor targeting framework that couples a high-level Future-State Visuomotor Target Predictor with a low-level Target-Conditioned Structured Dexterous Policy, separating coarse future-state guidance from high-frequency dexterous action execution.
\item We introduce an action-free future target prediction module that replaces privileged future demonstration targets with predicted targets conditioned on RGB and proprioceptive/geometric history, using structured visuomotor tokenization and sparse multi-horizon target prediction.
\item We validate DexFuture on challenging bimanual dexterous tool-use tasks, showing that predicted targets recover most of the privileged-oracle performance, strongly outperform no-target control, and execute substantially faster than DexWM-style CEM planning.
\end{itemize}
\section{Related Work}
\paragraph{Dexterous manipulation from demonstrations and targets.}
Learning dexterous manipulation directly from sparse task rewards is difficult due to the high DOF of hands and contact-rich dynamics. A common strategy is to leverage human demonstrations, retargeted references, or structured future targets to make policy learning tractable~\cite{dexpilot,dexmv,bidexhands,dexpoint,unidexgrasp,dextrack,maniptrans,dexmachina,dexmimicgen,omnigrasp,physgraph}. Large hand-object datasets and egocentric manipulation datasets further provide rich supervision for learning hand trajectories, object interactions, and task structure~\cite{dexycb,arctic,grab,hoi4d,oakink2,egodex}. These works show the importance of demonstration-guided or target-conditioned control, especially for high-DoF dexterous hands. However, the future reference or target used by the policy is often extracted from the demonstration trajectory and contains privileged future hand, tool, or object states that are unavailable at deployment. DexFuture aims to retain the performance of target-conditioned policies, while obviating demonstration targets by \emph{predicting} the future target from visuomotor history.

\paragraph{World models and action-conditioned planning.}
World models learn predictive representations for decision making and have been widely used with model-predictive control and trajectory optimization~\cite{dreamerv3,tdmpc,tdmpc2,mwm,mvmwm}. Recent predictive representation models and video world models further scale future prediction with latent objectives or controllable generative models~\cite{ijepa,vjepa,nwm,dit}. For dexterous manipulation, DexWM~\cite{dexwm} learns an action-conditioned world model from large-scale videos and uses the cross entropy method (CEM) to plan next actions~\cite{cem}. Such action-conditioned models are suitable for counterfactual rollouts, but online planning over high-dimensional dexterous action sequences is expensive and difficult to run at contact-control frequency. Our proposed method, DexFuture, takes a different route: it uses an action-free predictor to generate future-state targets directly, then delegates high-frequency action execution to a target-conditioned policy.
\section{Method}

\subsection{Hierarchical Problem Formulation}

\begin{figure*}[!t]
  \centering
  \includegraphics[width=\textwidth]{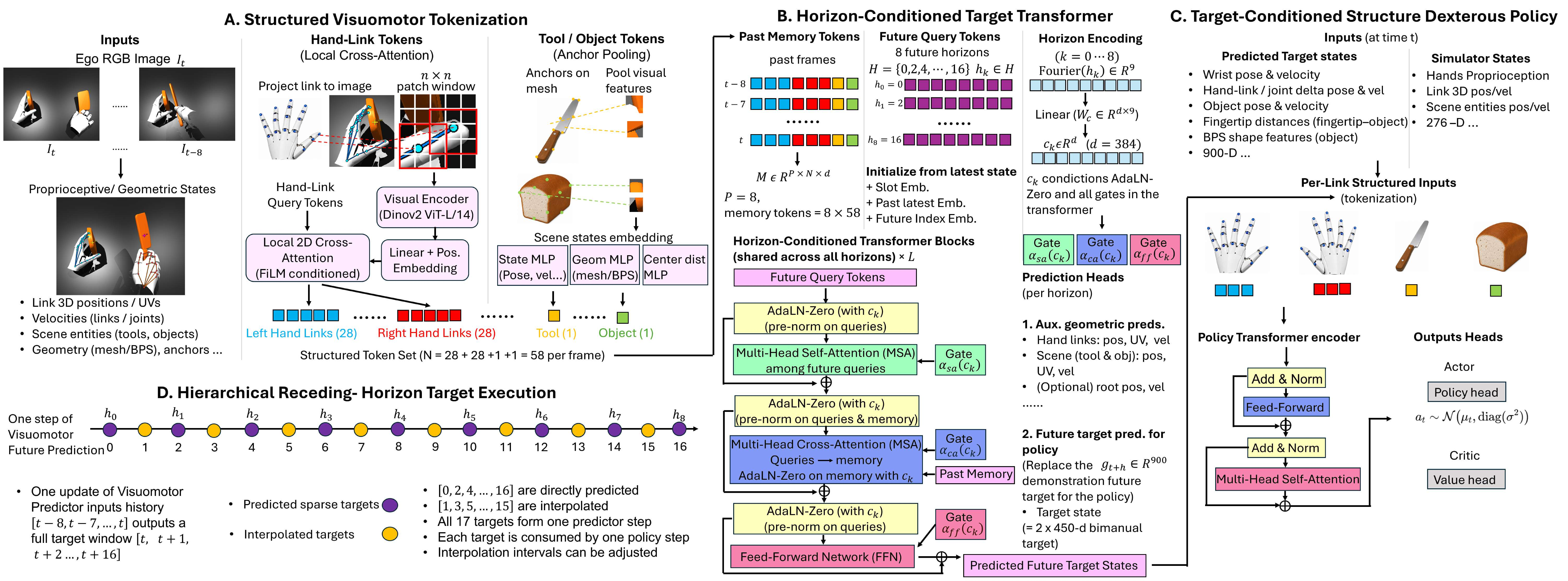}
\caption{
    \textbf{DexFuture overview.}
    DexFuture is a hierarchical system that separates bimanual dexterous manipulation into slow future target generation and fast action-level control.
    \textbf{A.} Given recent egocentric RGB observations and proprioceptive/geometric states, we construct structured visuomotor tokens instead of passing dense image patches to the predictor. Hand-link embeddings are obtained by projecting each link into the image and cross-attending to local visual patches, while tool/object embeddings are built by anchor-aligned visual sampling and entity-state/geometry embeddings.
    \textbf{B.} The Horizon-Conditioned Target Transformer takes the structured embedding history as memory and predicts sparse future structured embeddings at horizons $\mathcal{H}=\{0,2,4,\ldots,16\}$. Future embeddings are first initialized from the latest observed state and modulated with learned future-index embeddings and Fourier horizon encodings, then refined via self-attention and cross-attention to the visuomotor history using AdaLN-Zero transformer blocks. The predicted embeddings are decoded into auxiliary geometric states for supervision, and a $900$-D future target used by the low-level policy.
    \textbf{C.} The target-conditioned structured dexterous policy consumes the current state and the predicted target, tokenizes the bimanual hand-tool-object system into per-link and scene embeddings, and outputs a bimanual action distribution through a transformer encoder and policy head.
    \textbf{D.} During receding-horizon execution, a single forward pass of the visuomotor predictor produces a target trajectory over multiple time steps. Intermediate targets are linearly interpolated to allow high-frequency feedback control. The predictor is trained with supervised latent, state, and target losses, while the policy is trained with PPO against a tracking reward.
}
    \vspace{-5mm}
    \label{fig:policy_overview}
\end{figure*}




Figure \ref{fig:policy_overview} shows an overview of DexFuture. Let $s_t$ be the current robot-object state, $a_t$ be the bimanual hand action, and $g^{\mathrm{demo}}_{t+h}$ be a future demonstration target with horizon $h$. A target-conditioned dexterous policy can be written as
\begin{equation}
a_t \sim \pi_{\phi}(\cdot \mid s_t, g^{\mathrm{demo}}_{t+h}).
\label{eq:demo_target_policy}
\end{equation}
The target $g^{\mathrm{demo}}_{t+h}$ provides future reference for high-DoF dexterous control but it is privileged: at inference, the robot observes the current scene and proprioception, not the future demonstration state of the hands, tool, or object. 
Thus, our main objective is to predict future targets from the visuomotor history. We assume access to a history of $K$ RGB observations and proprioceptive/geometric cues:
\begin{equation}\label{eq:obs_history}
\mathcal{O}_{t-K:t}=\{I_{\tau},p_{\tau}\}_{\tau=t-K}^{t},
\end{equation}

where $I_{\tau}$ is the RGB observation and $p_{\tau}$ denotes structured proprioceptive and available geometric cues. Given this visuomotor history, DexFuture predicts future targets up to horizon $h$ as:
%
\begin{equation}\label{eq:future_target_prediction}
\hat{g}_{t+h}=F_{\theta}(\mathcal{O}_{t-K:t};\mathcal{H}).
\end{equation}
Here, $F_{\theta}$ is our \emph{Future-State Visuomotor Target Predictor}, and $\hat{g}_{t+h}$ is the predicted target trajectory, which has the same representation as the demonstrations $g^{\mathrm{demo}}_{t+h}$, but is inferred from observation history rather than being given from demonstrations. In practice, we predict targets up to a finite horizon $\mathcal{H}$ with coarser intervals, such as $\mathcal{H}=\{0,2,\ldots,16\}$, and use interpolation for intermediate control steps. The low-level policy then executes
\begin{equation}
a_{t+\delta}
\sim
\pi_{\phi}(\cdot \mid s_{t+\delta}, \tilde{g}_{t+\delta}),
\qquad
\tilde{g}_{t+\delta}
=
\mathrm{Interp}\left(\{\hat{g}_{t+h}\}_{h\in\mathcal{H}},\delta\right),
\label{eq:pred_target_policy}
\end{equation}

This formulation yields a compact hierarchy: $F_\theta$ predicts an action-free future target window over multiple low-level steps, while $\pi_\phi$ executes high-frequency actions using the interpolated targets. Unlike action-conditioned world-model planning, the predictor does not require candidate future actions as input, avoiding online action-sequence optimization and the circular dependency between future actions and future targets. The following sections describe how DexFuture constructs structured visuomotor embeddings, predicts sparse-horizon future targets, and executes them with the target-conditioned policy. We also provide more details and pseudo-code in the \emph{supplementary}.

\subsection{Future-State Visuomotor Target Predictor}



The \emph{Future-State Visuomotor Target Predictor} predicts the target states tracked by the low-level policy, rather than future pixels or actions. This follows the DexFuture hierarchy: the predictor provides coarse future hand-tool-object state guidance, while fine contact-rich control is handled by the low-level policy.

The target states exhibit a kinematic structure that allows decomposition into the hand, tool and the object. 
Therefore, it is unnecessary to use \emph{all} visual patches from the images to encode the visuomotor history, which may also be poorly aligned with the output.    
Instead, we propose a more structured embedding that exploits the physical relationship between the visual and proprioceptive information to extract only the relevant visual information.

For each observed frame, a frozen visual encoder extracts patch-level image features. Each hand link is projected into the image, and we collect the local visual neighborhood around the projected link. The query vector for the projected link, constructed from its identity, 3D position, and 2D projection, only attends to this local visual neighborhood to produce a link-level visuomotor embedding. Tool and object embeddings are constructed from their current state, geometry, type, projected center, and visual features. Together, these embeddings form a compact physical representation as
\begin{equation}
Z_t =
\{z^{\mathrm{hand}}_{t,i}\}_{i=1}^{N_h}
\cup
\{z^{\mathrm{tool}}_t,z^{\mathrm{obj}}_t\},
\label{eq:structured_tokens}
\end{equation}
where $Z_t$ is the structured embeddings at time $t$, $N_h$ is the number of hand-link tokens, $z^{\mathrm{hand}}_{t,i}$ is the embedding for hand link $i$, and $z^{\mathrm{tool}}_t,z^{\mathrm{obj}}_t$ are the tool and object embeddings. 

\paragraph{Horizon-Conditioned Target Transformer.}
Given the structured history $Z_{t-K:t}$, DexFuture predicts future targets at sparse timesteps over a horizon. The observed history embeddings serve as memory. Future query  vectors 
are initialized from the latest visuomotor embeddings 
$Z_t$, since future manipulation states are naturally predicted as transformations of the current hand-tool-object state. Each future timestep $h\in\mathcal{H}$ receives a horizon embedding, allowing the shared transformer to specialize its prediction for near and distant futures. We write this prediction compactly as
\begin{equation}
\hat{Z}_{t+h}=T_{\theta}(Z_t, Z_{t-K:t}, h),
\qquad h\in\mathcal{H},
\label{eq:horizon_transformer}
\end{equation}
where $T_{\theta}$ is the horizon-conditioned transformer and $\hat{Z}_{t+h}$ is the predicted structured embeddings at timestep $h$. Architecturally, this module uses adaptive layer-normalization conditioning from conditional diffusion transformer (CDiT) blocks \cite{cdit}. Horizon information modulates transformer updates through adaptive normalization and gated residual paths. However, unlike diffusion models, we remove the iterative denoising in DexFuture to directly regress sparse future targets for fast inference. The predicted future embeddings are decoded into auxiliary physical states and the target representation used by the dexterous policy:
\begin{equation}
\hat{g}_{t+h}=D_{\theta}(\hat{Z}_{t+h}),
\qquad h\in\mathcal{H},
\label{eq:target_decoder}
\end{equation}
where $D_{\theta}$ is the target decoder. The decoded target has the same layout as the demonstration target used by the policy so that the policy interface remains the same. During training, $F_{\theta}$ is supervised from demonstration using a future imitation target loss and an auxiliary future embedding loss: 
\begin{equation}
\mathcal{L}_{\mathrm{pred}}
=
\lambda_{\mathrm{state}}\mathcal{L}_{\mathrm{state}}
+
\lambda_{\mathrm{target}}\mathcal{L}_{\mathrm{target}}.
\label{eq:prediction_loss}
\end{equation}
Here, $\mathcal{L}_{\mathrm{state}}$ supervises future hand-link, tool, and object states, while $\mathcal{L}_{\mathrm{target}}$ supervises the decoded policy target. The exact target decomposition is described in the \textbf{supplementary material}.

\subsection{Target-Conditioned Structured Dexterous Policy}

DexFuture can be paired with any low-level dexterous controller that consumes a structured future target. In our experiments, we instantiate the controller as a target-conditioned per-link transformer policy, inspired from \cite{physgraph}. The policy tokenizes the current bimanual hand-tool-object state and the target into hand-link tokens, scene tokens, and a policy token. A transformer encoder processes these tokens, and the final policy token parameterizes a Gaussian action distribution,
\begin{equation}
a_t\sim\mathcal{N}\left(\mu_{\phi}(s_t,\tilde{g}_t),\Sigma_{\phi}\right),
\label{eq:policy_distribution}
\end{equation}
where $\mu_{\phi}$ is the action mean and $\Sigma_{\phi}$ is a learned diagonal covariance.

The policy is trained with PPO \cite{PPO} using tracking rewards against demonstration trajectories. The reward includes tracking terms for wrist pose, hand-link/fingertip positions, object pose, object velocity, object angular velocity, and energy regularization. During privileged baseline evaluation, 
the target is replaced with the demonstration target. 

\subsection{Hierarchical Receding-Horizon Execution}
At inference time, DexFuture executes the system hierarchy in a receding-horizon manner. At refresh time $t_j$, the predictor observes the recent history and produces a sparse future target sequence. For each low-level control step $t_j+\delta$ before the next refresh, the policy receives an interpolated target and computes an action from the current state:
\begin{equation}
\hat{\mathbf{g}}_{t_j:t_j+H}
=
F_{\theta}(\mathcal{O}_{t_j-K:t_j};\mathcal{H}),
\qquad
a_{t_j+\delta}
\sim
\pi_{\phi}(\cdot \mid s_{t_j+\delta},
\mathrm{Interp}(\hat{\mathbf{g}}_{t_j:t_j+H},\delta)).
\label{eq:receding_horizon_execution}
\end{equation}
This execution scheme enables the predictor to operate at a slower timescale while the policy remains reactive at the control timescale. The predicted targets need not be physically exact action-level rollouts. Instead, they provide coarser future state guidance, while the low-level policy performs finer contact-level correction and action execution.
\section{Experimental Results}
\label{sec:result}

\begin{table*}[t]
\centering
\caption{\textbf{Quantitative results on bimanual tool-use tasks of Oakink2 dataset \cite{oakink2}.} SR: Success Rate; E\_t: Tool \& Object Translation Error; E\_j: Hand Joint Error; E\_ft: Fingertip Error.}
\label{table:physgraph_results}
\resizebox{\linewidth}{!}{%
\begin{tabular}{lll cccc}
\toprule
\textbf{Task ID}  & \textbf{Task / Tool (R) / Object (L)} & \textbf{Methods} & \textbf{SR (\%)} $\uparrow$ & \textbf{E\_t (cm)} $\downarrow$ & \textbf{E\_j (cm)} $\downarrow$ & \textbf{E\_ft (cm)} $\downarrow$ \\ \midrule

\multirow{3}{*}{083f7@0}  & \multirow{3}{*}{cut / chop knife / bread} & Maniptrans (GT target) & 55.69 & 1.31 & 2.42 & 2.01 \\
 & & PhysGraph (GT target) & \textbf{90.05} & \textbf{0.69} & 2.17 & \textbf{1.42} \\
 & & PhysGraph (No target) & 4.16 & 1.87 & 3.11 & 2.68 \\
 & & DexFuture (Pred target) & 83.49 & 1.06 & \textbf{2.08} & 2.04 \\ \hline
 
 \multirow{3}{*}{9fc3e@0}  & \multirow{3}{*}{cut / fruit knife / apple} & Maniptrans (GT target) & 70.58 & 0.84 & 2.99 & \textbf{1.97} \\
 & & PhysGraph (GT target) & 87.87 & 0.98 & 2.04 & 2.17 \\ 
 & & PhysGraph (No target) & 20.5 & 1.42 & 2.34 & 2.45 \\
 & & DexFuture (Pred target) & \textbf{89.79} & \textbf{0.61} & \textbf{2.03} & 2.40 \\ \hline

\multirow{3}{*}{1292e@0} & \multirow{3}{*}{pour / mug / mug} & Maniptrans (GT target) & 45.60 & 6.78 & 3.50 & \textbf{3.35} \\
 & & PhysGraph (GT target) & \textbf{49.77} & \textbf{5.93} & \textbf{3.07} & 4.33 \\
 & & PhysGraph (No target) & 1.82 & 8.37 & 4.13 & 4.48 \\
 & & DexFuture (Pred target) & 41.05 & 7.27 & 3.82 & 3.85 \\ \hline

\multirow{3}{*}{817fb@0} & \multirow{3}{*}{wipe / big brush / whiteboard} & Maniptrans (GT target) & 50.05 & 1.55 & 2.17 & 2.24 \\
 & & PhysGraph (GT target) & \textbf{62.24} & \textbf{1.33} & \textbf{1.54} & \textbf{1.86} \\
 & & PhysGraph (No target) & 8.99 & 2.77 & 2.68 & 2.58 \\
 & & DexFuture (Pred target) & 56.96 & 1.50 & 2.42 & 2.43 \\ \hline

 \multirow{3}{*}{fc88d@0} & \multirow{3}{*}{wipe / small brush/ whiteboard} & Maniptrans (GT target) & 71.21 & 1.21 & \textbf{2.44} & 1.79 \\
 & & PhysGraph (GT target) & \textbf{80.65} & \textbf{1.11} & 2.49 & \textbf{1.50} \\
 & & PhysGraph (No target) & 4.87 & 2.23 & 3.72 & 3.04 \\
 & & DexFuture (Pred target) & 67.17 & 1.25 & 3.02 & 1.70 \\ \hline

\multirow{3}{*}{e1fa6@0} & \multirow{3}{*}{shear / scissors / paper} & Maniptrans (GT target) & 15.16 & \textbf{1.18} & 3.31 & 2.25 \\
 & & PhysGraph (GT target) & \textbf{35.84} & 1.35 & \textbf{2.52} & \textbf{1.83} \\
 & & PhysGraph (No target) & 0.00 & 1.78 & 3.70 & 2.68 \\
 & & DexFuture (Pred target) & 30.69 & 1.62 & 2.97 & 1.92 \\ \hline

 \multirow{3}{*}{9bb17@5} & \multirow{3}{*}{shear / scissors / paper} & Maniptrans (GT target) & 51.68 & 1.93 & \textbf{2.26} & 2.86 \\
 & & PhysGraph (GT target) & \textbf{59.21} & \textbf{1.90} & 3.12 & \textbf{2.70} \\
 & & PhysGraph (No target) & 11.18 & 3.85 & 3.81 & 3.93 \\
 & & DexFuture (Pred target) & 48.66 & 2.49 & 2.73 & 2.90 \\

\bottomrule
\vspace{-3mm}
\label{tab:policy}
\end{tabular}}
\end{table*}

\subsection{Experimental Setup}

We evaluate DexFuture on challenging bimanual dexterous tool-use tasks from OakInk2 \cite{oakink2}, including cutting, pouring, wiping, shearing, and stirring. Each task requires coordinated interaction among two dexterous hands, a tool, and an object. We compare DexFuture against two strong target-conditioned dexterous policy baselines: ManipTrans \cite{maniptrans} and PhysGraph \cite{physgraph}. For these baselines, the target is provided from the \emph{ground-truth} future demonstration state. We also include a no-target variant of the PhysGraph policy, which removes the future target input and tests whether the policy can perform the task from the current state alone. We report success rate (SR), tool/object translation error $E_t$, hand joint error $E_j$, and fingertip error $E_{ft}$. Higher SR is better, while lower tracking errors are better. Implementation details, reward terms, training hyperparameters, evaluation metrics definition, and results of more tasks are provided in the \textbf{supplementary}.

\subsection{Policy Evaluation with Predicted Future Targets}

The core question of policy evaluation is whether DexFuture can retain the benefit of target-conditioned dexterous control without using privileged future demonstration targets at inference. Table~\ref{tab:policy} summarizes the policy performance. The ``no-target'' policy shows that simply removing the future target makes high-DoF dexterous tool use nearly infeasible, while DexFuture recovers a strong target signal from visuomotor history.

Compared with the privileged structured-policy, DexFuture achieves about $90\%$ of the average success rate of privileged baselines ($59.69\%$ vs. $66.52\%$), despite using predicted targets instead of ground-truth future demonstration targets. On the fruit-knife cutting task, DexFuture even slightly outperforms the privileged baseline in success rate ($89.79\%$ vs. $87.87\%$) and tool/object translation error ($0.61$cm vs. $0.98$cm). On the bread-cutting, whiteboard-wiping, and paper-shearing tasks, DexFuture remains close to the privileged baseline while significantly outperforming the no-target variant. These results support our main claim: future target guidance is essential for dexterous tool use, but it does not have to be provided by online privileged demonstration state.


\subsection{Future-State Visuomotor Target Prediction}
\label{sec:exp:predictor}

\begin{wraptable}[12]{r}{0.50\textwidth}
\vspace{-0.8em}
\centering
\caption{\textbf{Future-state target prediction accuracy.}
We report 3D error, UV error, and PCK within 5/10 pixels.}
\label{tab:predictor_eval}
\vspace{-0.5em}
\footnotesize
\setlength{\tabcolsep}{2pt}
\begin{tabular*}{\linewidth}{@{\extracolsep{\fill}}lcccc@{}}
\toprule
\textbf{Task ID} &
\textbf{3D} $\downarrow$ &
\textbf{UV} $\downarrow$ &
\textbf{PCK@5} $\uparrow$ &
\textbf{PCK@10} $\uparrow$ \\
\midrule
083f7@0  & 1.47 & 3.32  & 79.56 & 98.37 \\
9fc3e@0  & 0.87 & 2.56  & 91.34 & 99.78 \\
598a5@0  & 1.31 & 2.73  & 90.24 & 97.82 \\
1292e@0  & 2.31 & 6.91  & 38.49 & 80.02 \\
817fb@0  & 4.54 & 8.74  & 19.33 & 68.27 \\
e1fa67@0 & 5.54 & 25.58 & 7.98  & 32.93 \\
9bb17@5  & 4.50 & 12.86 & 22.31 & 54.15 \\
cde36@1  & 1.65 & 5.41  & 76.93 & 95.76 \\
\bottomrule
\end{tabular*}
\vspace{-0.9em}
\end{wraptable}

We next evaluate the Future-State Visuomotor Target Predictor independently from the policy. Table~\ref{tab:predictor_eval} reports future-state prediction quality with 3D error in cm, 2D UV error in pixels, and PCK (Percentage of Correct Keypoints) in 5 and 10 pixels. The predictor performs accurately on cutting/stirring tasks, while shearing is more challenging. 
The performance of the target predictor shows a consistent pattern with that of the overall policy. Challenging tasks involve narrow tool-object contact regions, and abrupt motion changes, making future states harder to infer from the observation history alone. The tasks considered for testing are all unseen during training. We provide the full training tasks ID in the \textbf{Supplementary Material}.

\begin{wraptable}[11]{r}{0.50\linewidth}
\vspace{-1.0em}
\centering
\caption{\textbf{Ablation on future prediction horizon.}
We compare different future horizon schedules.}
\label{tab:horizon_ablation}
\vspace{-0.5em}
\resizebox{\linewidth}{!}{%
\begin{tabular}{lcccc}
\toprule
\textbf{Task ID} &
\textbf{3D} $\downarrow$ &
\textbf{UV} $\downarrow$ &
\textbf{PCK@5} $\uparrow$ &
\textbf{PCK@10} $\uparrow$ \\
\midrule
083f7@0\_h24  & 1.66 & 4.05 & 71.87 & 97.54 \\
083f7@0\_h16  & 1.47 & 3.32 & 79.56 & 98.37 \\
083f7@0\_h8   & 1.19 & 3.68 & 76.99 & 96.34 \\
9fc3e@0\_h24  & 1.11 & 2.95 & 86.66 & 99.27 \\
9fc3e@0\_h16  & 0.87 & 2.56 & 91.34 & 99.78 \\
9fc3e@0\_h8   & 0.84 & 2.46 & 92.15 & 99.72 \\
598a5@0\_h24  & 1.77 & 3.97 & 73.56 & 97.46 \\
598a5@0\_h16  & 1.31 & 2.73 & 90.24 & 97.82 \\
598a5@0\_h8   & 1.38 & 2.86 & 89.72 & 97.40 \\
\bottomrule
\end{tabular}}
\vspace{-1.0em}
\end{wraptable}

Table~\ref{tab:horizon_ablation} shows the effect of the receding horizon, which is directly tied to the temporal hierarchy in DexFuture. A short horizon is easier to predict, but provides less future guidance to the controller. A long horizon covers more policy steps, but becomes harder to predict reliably. We choose $h16$ as the default horizon, corresponding to $\{0,2,4,\ldots,16\}$, which provides the best overall trade-off. This supports our hierarchical design that the high-level predictor produces future targets at a slower timescale, while the policy executes dense low-level actions.

\subsection{Qualitative Results}

\begin{figure*}[!t]
  \centering
  \includegraphics[width=\textwidth]{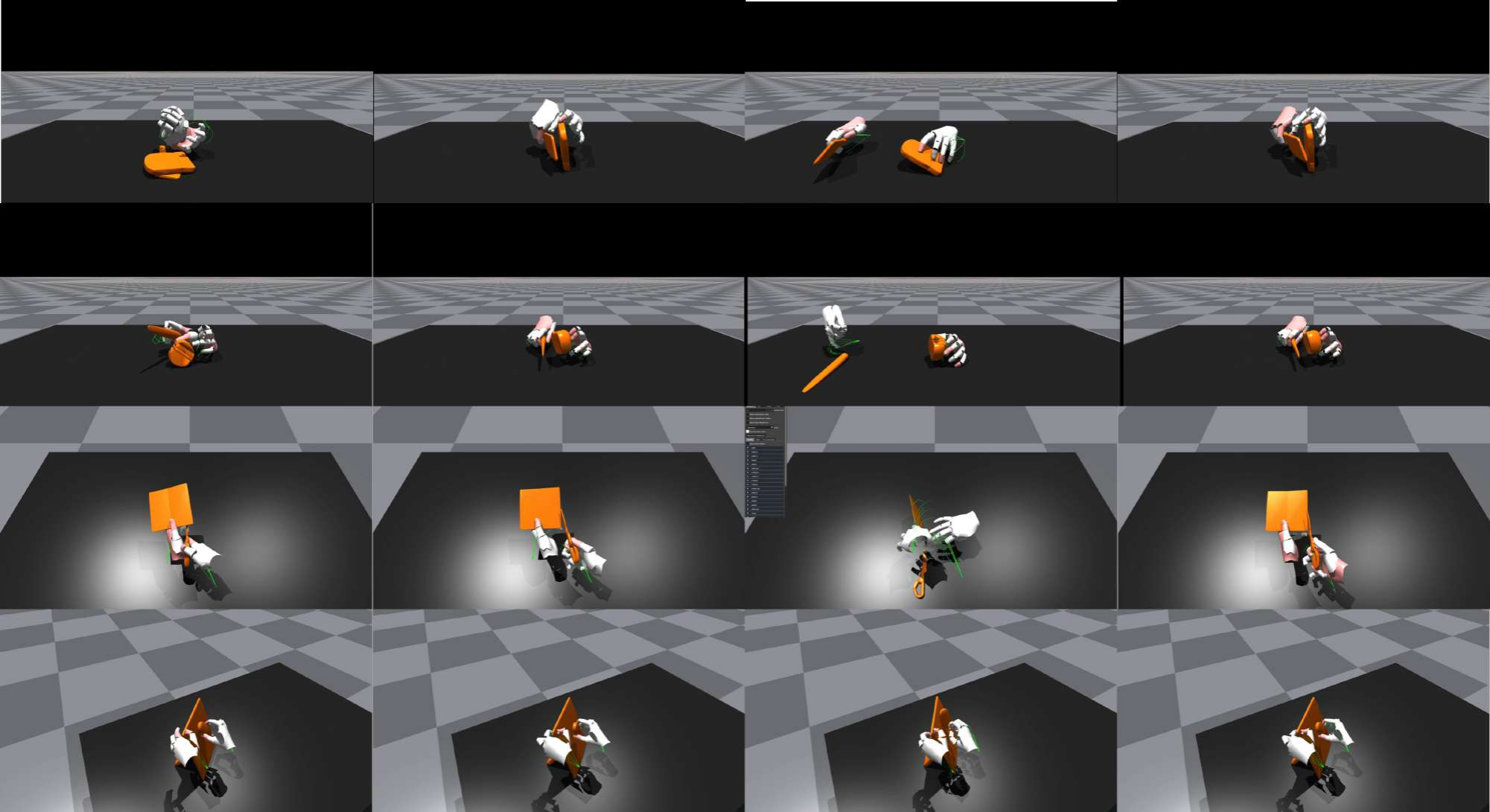}
\caption{\textbf{Qualitative comparison on bimanual dexterous tool-use tasks.}
From top to bottom: 1. chop knife cutting bread (083f7@0). 2. fruite knife cutting apple (9fc3e@0). 3. shear paper with scissors (9bb17@5). 4. wipe whiteboard with brush (fc88d@0). From left to right: 1. ManipTrans with ground-truth target. 2. PhysGraph with ground-truth target 3. PhysGraph without target. 4. Our DexFuture with predicted target. The privileged baselines can access the demonstration targets, while DexFuture infers the future target from visuomotor history alone. The no-target policy fails to maintain robust hand-tool-object interaction, whereas DexFuture recovers stable tool use and produces rollouts comparable to the oracle policies. We include more tasks and full videos in the \textbf{Supplementary Material}.
}
\label{fig:results}
\end{figure*}

Figure~\ref{fig:results} visualizes representative rollouts. The no-target policy often loses stable hand-tool-object coordination, confirming that current-state feedback alone is insufficient for these tasks. In contrast, DexFuture produces motions that closely follow the oracle target-conditioned policies: the hands maintain tool grasp, approach the object with appropriate alignment, and complete contact-rich interactions such as cutting, wiping, and shearing. These qualitative results complement Table~\ref{tab:policy}, which shows that DexFuture is not simply improving a scalar success metric, but restoring the coordinated structure that target-conditioned dexterous policies rely on. The predicted target acts as a coarse future guidance, while the policy performs the local contact correction for execution.

\subsection{Comparison with Action-Conditioned Planning}
\label{sec:dexwm}
To validate our choice of \emph{action-free} predictor with a policy rather than an \emph{action-conditioned} world model with planning for dexterous tool use, we further compare DexFuture with the strongest action-conditioned dexterous world-model planning baseline following DexWM~\cite{dexwm}. We train the default DexWM model on the EgoDex dataset \cite{egodex}, evaluate on DexWM's default dataset to get comparable results to those reported in \cite{dexwm}, then further evaluate variants with and without finetuning on our OakInk2 tool-use tasks. We test both short and long-horizon planning. The best performance was obtained by finetuning on the OakInk2 dataset, and by using a planning horizon of 1 (next RGB frame as the goal). However, it still underperforms DexFuture. More importantly, testing using the same single 3090Ti GPU, DexWM's default CEM-based planning runs at approximately $0.24$Hz, while DexFuture executes at the policy control rate of $60$Hz. The $250\times$ speed gap is critical for high-DoF dexterous manipulation, where contact-rich actions must be updated at high frequency. These results highlight the practical advantage of our hierarchy: instead of optimizing over counterfactual future action sequences online, DexFuture amortizes future target generation into a low-frequency visuomotor predictor and leaves high-frequency action control to the policy. The full comparison is described in the \textbf{Supplementary Material}.

\section{Conclusion}
\label{sec:conclusion}
We presented \textbf{DexFuture}, a hierarchical future-state visuomotor targeting framework for bimanual dexterous tool use. DexFuture removes the need for privileged future demonstration targets by predicting future targets conditioned on RGB and proprioceptive/geometric history, while a low-level structured policy executes high-frequency contact-rich actions. This design preserves the benefit of target-conditioned dexterous control without requiring future demonstration states or slow online action-sequence planning. Experiments on challenging tool-use tasks show that DexFuture mostly retains the performance of privileged baselines, strongly outperforms no-target control, and runs substantially faster than counterfactual planning.
\section{Limitations and Future Work}
\label{sec:limitations}
The main remaining bottleneck is future target prediction under difficult contacts. Tasks with narrow contact regions or abrupt tool motion remain challenging. The high-level predictor must be robust enough to guide the precise action of low-level policy. Potential future solutions should focus on uncertainty-aware or contact-aware prediction, stronger visual grounding under occlusion, and deployment to a real robot.

\clearpage

\bibliography{example}  

\clearpage
\section{Appendix}
\subsection{Comparison with Action-Conditioned Planning}
In this section, we complete the comparison in Section \ref{sec:dexwm}. We compare DexFuture against an action-conditioned world-model planning baseline following DexWM ~\cite{dexwm}. The original DexWM planner performs image-goal CEM: at each MPC step, a goal RGB image is encoded into a DexWM visual latent, candidate future robot trajectories are sampled, and each candidate is rolled out through the action-conditioned world model. Candidates are scored by the MSE between the predicted visual latent and the goal-image latent. The planner then refits the CEM distribution using the elite samples and executes the first action before replanning.

In the original DexWM CEM baseline, their CEM-planning configuration is highly expensive supported by 8 H100 GPUs for inference: prediction horizon is 3, CEM optimization steps is 10, candidate samples per CEM iteration is 1024, and elite samples are 10 for distribution refitting. This setting is computationally heavy because every CEM iteration requires thousands of autoregressive visual world-model rollouts.

To make the baseline feasible to run, we implement a state-based CEM variant. Instead of scoring candidates by image-latent distance to a goal RGB frame, we score predicted future states directly in the target state space used by the controller. The state-based CEM uses a smaller online budget: horizon 16, 128 samples, 16 elites, and 4 CEM iterations, with micro-batched mixed-precision scoring. This reduces memory and compute while aligning the planning objective with the downstream policy.  On our single 3090Ti evaluation GPU, this CEM-based planning runs at around 0.24 Hz, far below the control frequency needed for contact-rich dexterous manipulation. Despite these adaptations, DexWM-style online planning remains substantially slower than DexFuture, which amortizes future target prediction into System-1 and executes System-0 at the policy control rate of 60 Hz.

We evaluate four DexWM-style baselines by varying two factors: the world-model checkpoint and the planning horizon. For the checkpoint, following DexWM's default training, we first train on EgoDex \cite{egodex} dataset and test the original checkpoint to get the equivalent results as DexWM's paper reported, then we finetune DexWM on OakInk2 \cite{oakink2}. For the horizon, we evaluate a short-horizon oracle setting with horizon 1, where the planner is given the next-frame ground-truth goal at every step, and a longer-horizon setting with horizon 16. In the image-goal DexWM formulation, the horizon-1 setting corresponds to using the next RGB frame as the goal, while in our state-based adaptation, the analogous setting uses the next ground-truth target state.

\begin{figure}[!htbp]
  \centering
  \includegraphics[width=\textwidth]{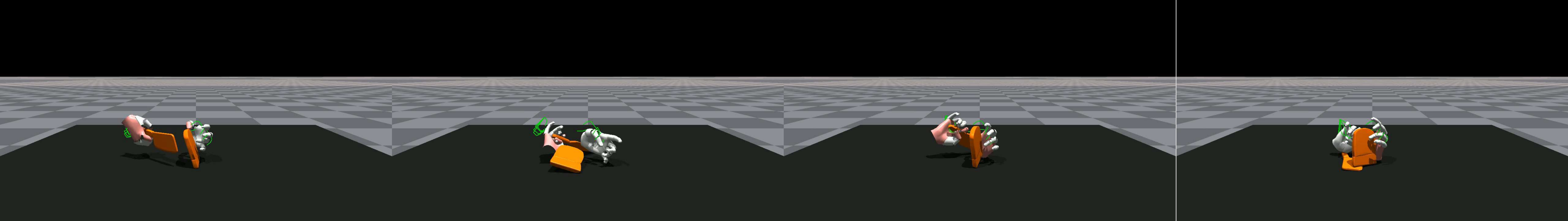}
\caption{\textbf{Qualitative results of DexWM-style CEM-based Planning.}
From left to right: Finetune + Horizon 1; No-finetune + Horizon 1; Finetune + Horizon 16; No-finetune + Horizon 16. 
}
\label{fig:planning}
\end{figure}

\begin{table}[!htbp]
\centering
\caption{\textbf{Quantitative results of DexWM-style CEM-based Planning.} Speed is tested on the same single 3090Ti GPU. SR: Success Rate; E\_t: Tool \& Object Translation Error; E\_j: Hand Joint Error; E\_ft: Fingertip Error.}
\label{table:physgraph_results}
\resizebox{\linewidth}{!}{%
\begin{tabular}{l ccccc}
\toprule
\textbf{Ablations} & \textbf{Speed (Hz)} $\uparrow$ & \textbf{SR (\%)} $\uparrow$ & \textbf{E\_t (cm)} $\downarrow$ & \textbf{E\_j (cm)} $\downarrow$ & \textbf{E\_ft (cm)} $\downarrow$ \\ \midrule
 DexWM (GT target, Finetune + Horizon 1) & 0.24 & 69.57 & 1.20 & 2.29 & 1.93 \\
 DexWM (GT target, No-finetune + Horizon 1) & 0.24 & 0 & 2.94 & 3.65 & 3.41 \\
 DexWM (GT target, Finetune + Horizon 16) & 0.24 & 0 & 2.82 & 2.78 & 2.55 \\
 DexWM (GT target, No-finetune + Horizon 16) & 0.24 & 0 & 3.17 & 2.97 & 2.92 \\
 DexFuture (Pred target, Horizon 16) & \textbf{60} & \textbf{83.49} & \textbf{1.06} & \textbf{2.08} & \textbf{2.04} \\
 
\bottomrule
\label{tab:planning}
\end{tabular}}
\end{table}

Figure \ref{fig:planning} and table \ref{tab:planning} indicate that dense goals and finetuning are both essential for the world model planning. With the per-step densest ground truth guidance, DexWM achieves impressive performance. However, the ground truth targets are not always available during the inference, the rollouts sampling also require significant longer time than a well-trained policy, further diminishing the practical efficiency of a future action-conditioned world model on high-DoF dexterous tasks. By comparison, our proposed Please check the full \textbf{video} comparison for more details.

\subsection{Relation to Action Chunking and Direct Action Prediction}
\label{supp:action_chunking}

DexFuture is related to action-chunking and diffusion-based visuomotor policies since all of these methods reason over a short future horizon. However, the predicted quantity is fundamentally different. Action-chunking methods such as ACT directly generate future motor commands, while diffusion or flow-based policies model the distributions over future action trajectories.
Our point is that direct action generation couples two problems that DexFuture separates: predicting where the hand-tool-object system should progress, and deciding how to achieve that progress through contact-rich actions.

This distinction is especially important for high-DoF dexterous hands. In an action chunk, the temporal plan and the low-level motor command are both encoded in action space. If contact occurs earlier or later than predicted, or if the tool-object alignment changes slightly, the remaining action chunk becomes unrobust. Receding-horizon execution can mitigate this by querying the policy repeatedly, but the policy still has to learn long-horizon task progress and fine contact correction through direct action prediction. DexFuture instead predicts a future target-state sequence, not a motor-command sequence. The low-level policy observes the current simulator state at every step and converts the current predicted target into an action, allowing contact correction.

Therefore, DexFuture should not be interpreted as an action-chunking method. Action-chunking predicts future actions; DexFuture predicts future target states. Action chunks are executed directly or through receding-horizon aggregation; DexFuture targets are interpreted by a separate policy. This separation lets the high-level module focus on coarse future-state guidance, while the low-level controller remains responsible for high-frequency dexterous contact execution.

DexFuture is also different from large VLA or diffusion foundation policies. These models generally map observations, and optionally language, to actions or action chunks. DexFuture instead assumes a target-conditioned dexterous controller and studies how to produce its future target input without privileged demonstration state. These directions are complementary: a direct-action policy or VLA backbone could potentially serve as the low-level controller, while DexFuture's target predictor provides structured long-horizon guidance.

\subsection{Method and Implementation Details}
\subsubsection{Observation and Target Notation}
\label{supp:obs_target_notation}

Let a demonstration trajectory be denoted by
\begin{equation}
\tau=\{I_t,p_t,s_t,g^{\mathrm{demo}}_t\}_{t=1}^{T},
\end{equation}
where $I_t$ is the egocentric RGB observation, $p_t$ contains structured proprioceptive and geometric cues available to the predictor, $s_t$ is the simulator state used by the policy, and $g^{\mathrm{demo}}_t$ is the demonstration target consumed by the target-conditioned dexterous policy.

The Future-State Visuomotor Target Predictor receives a history window
\begin{equation}
\mathcal{O}_{t-K:t}=\{I_\tau,p_\tau\}_{\tau=t-K}^{t},
\end{equation}
and predicts targets over a finite horizon set
\begin{equation}
\mathcal{H}=\{h_1,\ldots,h_M\}.
\end{equation}
In our default setting, $K=8$, so the predictor observes $9$ frames, and $\mathcal{H}=\{0,2,4,\ldots,16\}$. We additionally evaluate alternative horizon schedules and show results in Table \ref{tab:horizon_ablation}.

The structured state contains $N_\ell$ hand-link entries and two scene-level entries for the tool and object. Our model supports multiple tools and objects, where the multi entities in the environment are pooled to a tool entry and an object entry. This design facilitates the tasks where multi objects and tools are involved in the tool use. In our implementation, $N_\ell=56$ for the two hands, and the resulting structured token set has $N=N_\ell+2=58$ tokens per frame. The target $g_t$ has the same semantic layout as the policy's original demonstration target, so the downstream controller can consume either $g^{\mathrm{demo}}_t$ or the predicted target $\hat{g}_t$ without changing the policy interface.

\subsubsection{Structured Visuomotor Tokenization}
\label{supp:structured_tokenization}

The predictor does not run future prediction over all dense image patches. Instead, it converts each observation frame into a compact set of physical tokens corresponding to hand links, the tool, and the object.

Let $\Psi$ be a frozen visual encoder. For each frame $I_t$, the visual encoder produces patch features
\begin{equation}
V^{\mathrm{raw}}_t=\Psi(I_t)\in\mathbb{R}^{P\times d_v},
\end{equation}
where $P$ is the number of image patches and $d_v$ is the raw visual feature dimension. A learned projection maps these features into the structured token space,
\begin{equation}
V_t = W_v V^{\mathrm{raw}}_t + E_{\mathrm{patch}},
\qquad
V_t\in\mathbb{R}^{P\times d}.
\end{equation}
Here, $d$ is the structured token dimension and $E_{\mathrm{patch}}$ is a learned patch-position embedding. In our implementation, the frozen visual encoder is DINOv2 ViT-L/14, $d_v=1024$, and $d=256$. The projection is important because hand-link queries, tool/object descriptors, and future tokens are all represented in the same structured token space.

\paragraph{Hand-link tokens.}
For each hand link $\ell$, let $x_{t,\ell}\in\mathbb{R}^{3}$ be its 3D position, $u_{t,\ell}\in\mathbb{R}^{2}$ be its projected image coordinate, and $\dot{x}_{t,\ell}\in\mathbb{R}^{3}$ be its finite-difference velocity. We define a link-conditioned geometric feature
\begin{equation}
\xi_{t,\ell}
=
[\dot{x}_{t,\ell},\gamma_x(x_{t,\ell}),\gamma_u(u_{t,\ell})],
\end{equation}
where $\gamma_x$ and $\gamma_u$ are Fourier feature embeddings for 3D and 2D coordinates. Each link also has a learned identity embedding $e_\ell$. The base link query is
\begin{equation}
q^{0}_{t,\ell}
=
\mathrm{LN}\left(
W_{\mathrm{id}} e_\ell
+
W_{\xi}\xi_{t,\ell}
\right).
\end{equation}
The identity embedding tells the model which physical link is being queried, while $\xi_{t,\ell}$ tells the model where that link is, where it projects in the image, and how it is moving. However, a fixed additive query is still limited: the same link identity should attend to image evidence differently when it is near a tool, far from the object, or moving quickly during contact. We therefore use feature-wise linear modulation (FiLM) to adapt the query according to the current link geometry and motion:
\begin{equation}
(\alpha_{t,\ell},\beta_{t,\ell}) = f_{\mathrm{film}}(\xi_{t,\ell}),
\qquad
q_{t,\ell}=(1+\alpha_{t,\ell})\odot q^{0}_{t,\ell}+\beta_{t,\ell}.
\end{equation}
Conditioning FiLM on $\xi_{t,\ell}$ makes the query geometry-dependent while preserving the link identity. For example, the same fingertip token can produce different attention queries depending on whether it is approaching the tool, already in contact, or moving away from the object.

Around the projected coordinate $u_{t,\ell}$, we gather a local patch neighborhood $\Omega(u_{t,\ell})$. The hand-link token is computed by cross-attention from the link query to the local visual patch tokens:
\begin{equation}
z^{\mathrm{hand}}_{t,\ell}
=
q_{t,\ell}
+
\mathrm{MHA}
\left(
q_{t,\ell},
\{\mathcal{V}_{t,i}+\rho(\bar{u}_{i}-u_{t,\ell})\}_{i\in\Omega(u_{t,\ell})}
\right),
\end{equation}
where $\rho(\cdot)$ encodes relative 2D offsets between the queried link and the local patch centers. In our implementation, $\Omega(\cdot)$ is a $5\times5$ patch window. Thus each hand-link token is a local visual-geometric descriptor grounded at a physical link.

\paragraph{Tool and object tokens.}
Tool and object tokens are constructed from scene entities. To support scenes with different numbers of tools or objects, we allocate a fixed maximum number of entity slots $E_{\max}$ and use a binary validity mask $\omega_{t,e}\in\{0,1\}$ for each slot. If a scene contains fewer than $E_{\max}$ entities, unused slots are masked out and do not contribute to pooling. If a scene contains more than $E_{\max}$ entities, the current implementation keeps the first $E_{\max}$ entity specifications.

Let $e\in\{1,\ldots,E_{\max}\}$ index an entity slot, with state $r_{t,e}$, static geometry descriptor $m_e$, type label $c_e$, projected center $(x^c_{t,e},u^c_{t,e})$, and a set of anchors $\{(x^a_{t,e,k},u^a_{t,e,k})\}_{k=1}^{A}$. The state $r_{t,e}$ contains position, orientation, linear velocity, and angular velocity. The geometry descriptor $m_e$ is static for each entity, while the center and anchor locations are transformed and projected according to the current entity pose.

Each anchor samples a visual patch feature from the projected image location. We define
\begin{equation}
a_{t,e,k}
=
W_a[x^a_{t,e,k},u^a_{t,e,k}]
+
W_{\mathrm{vis}}\mathcal{V}_{t,\mathrm{sample}(u^a_{t,e,k})}.
\end{equation}
Anchor features are averaged over valid anchors:
\begin{equation}
\bar{a}_{t,e}=\frac{1}{A_e}\sum_{k=1}^{A_e}a_{t,e,k}.
\end{equation}
The entity token is
\begin{equation}
z^{\mathrm{ent}}_{t,e}
=
\mathrm{LN}
\left(
E_{\mathrm{type}}(c_e)
+
W_r r_{t,e}
+
W_m m_e
+
W_c[x^c_{t,e},u^c_{t,e}]
+
\bar{a}_{t,e}
\right).
\end{equation}

The final scene tokens are obtained by type-masked pooling. Let
\begin{equation}
\mathcal{E}^{\mathrm{tool}}_t
=
\{e:\omega_{t,e}=1,\ c_e=\mathrm{tool}\},
\qquad
\mathcal{E}^{\mathrm{obj}}_t
=
\{e:\omega_{t,e}=1,\ c_e=\mathrm{object}\}.
\end{equation}
Then
\begin{equation}
z^{\mathrm{tool}}_t
=
\frac{1}{|\mathcal{E}^{\mathrm{tool}}_t|}
\sum_{e\in\mathcal{E}^{\mathrm{tool}}_t}
z^{\mathrm{ent}}_{t,e},
\qquad
z^{\mathrm{obj}}_t
=
\frac{1}{|\mathcal{E}^{\mathrm{obj}}_t|}
\sum_{e\in\mathcal{E}^{\mathrm{obj}}_t}
z^{\mathrm{ent}}_{t,e}.
\end{equation}
Thus, although each frame may contain a variable number of valid scene entities, the predictor receives a fixed-size scene representation with one tool token and one object token per frame.

Unlike hand link tokens using local cross-attention over a patch window, we use such anchor-aligned visual sampling and pooling for scene (object/tool) tokens. This is sufficient because tool/object entities are spatially larger and already have explicit state, geometry, center, and multi-anchor information, while hand links are small and benefit more from local visual attention. In our implementation, anchor points are selected deterministically from the entity mesh vertices using evenly spaced vertex indices, transformed by the current entity pose, and projected into the egocentric image. This gives each scene token access to multiple local visual regions rather than only the object center.

\subsubsection{Horizon-Conditioned Target Transformer}
\label{supp:horizon_transformer}

Given history tokens $Z_{t-K:t}$, the predictor estimates future structured tokens $\hat{Z}_{t+h}$ for each $h\in\mathcal{H}$. We first project structured tokens into a transformer hidden space:
\begin{equation}
X_{\tau}=W_{\mathrm{in}}Z_{\tau},
\qquad \tau\in[t-K,t].
\end{equation}
The observed tokens form the memory
\begin{equation}
M=\mathrm{Flatten}(X_{t-K:t}),
\end{equation}
where flattening is over time and token index.

For each horizon $h_j\in\mathcal{H}$, where $j$ indexes the output horizon slot, the future query tokens are initialized from the latest observed structured state. Let $i\in\{1,\ldots,N\}$ denote the structured token index, corresponding to a hand link, tool token, or object token. We initialize
\begin{equation}
Y^0_{h_j,i}
=
X_{t,i}
+
E_{\mathrm{slot}}(i)
+
E_{\mathrm{frame}}(j),
\end{equation}
where $E_{\mathrm{slot}}(i)$ is a learned embedding for the physical token slot and preserves whether the query corresponds to a specific hand link, the tool, or the object. $E_{\mathrm{frame}}(j)$ is a learned embedding for the discrete future output slot. The slot embedding provides token identity, while the frame embedding distinguishes different predicted slots in the output sequence.

The numeric prediction horizon is encoded separately by Fourier features:
\begin{equation}
c_{h_j}=f_h(\gamma(h_j)),
\end{equation}
where $\gamma(h_j)$ is the Fourier encoding of the actual future offset $h_j$. This is different from only using a learned output-slot embedding: the same output slot can correspond to different numeric horizons under different horizon schedules, while $c_{h_j}$ explicitly tells the model the actual future offset.

Each transformer block uses horizon-conditioned adaptive normalization. For a token sequence $Y_h$ at horizon $h$, we define
\begin{equation}
\mathrm{AdaLN}(Y_h,c_h)
=
\mathrm{LN}(Y_h)\odot(1+s(c_h))+b(c_h),
\end{equation}
where $s(c_h)$ and $b(c_h)$ are horizon-conditioned scale and shift. One block updates the future queries as
\begin{align}
Y_h
&\leftarrow
Y_h
+
\alpha_{\mathrm{sa}}(c_h)\,
\mathrm{MSA}(\mathrm{AdaLN}(Y_h,c_h)),
\\
Y_h
&\leftarrow
Y_h
+
\alpha_{\mathrm{ca}}(c_h)\,
\mathrm{MCA}(\mathrm{AdaLN}(Y_h,c_h),M),
\\
Y_h
&\leftarrow
Y_h
+
\alpha_{\mathrm{ff}}(c_h)\,
\mathrm{FFN}(\mathrm{AdaLN}(Y_h,c_h)).
\end{align}
Here, MSA is self-attention among predicted future tokens, MCA is cross-attention from future queries to observed memory tokens, and FFN is the feed-forward network. The gates $\alpha_{\mathrm{sa}}$, $\alpha_{\mathrm{ca}}$, and $\alpha_{\mathrm{ff}}$ are also functions of the horizon condition. The adaptive conditioning is inspired from CDiT-style transformers \cite{cdit}, but our model is not a diffusion model: it has no noise injection, denoising objective, reverse sampling chain, or stochastic generation process. It directly regresses future structured tokens.

After $L$ blocks, we project back to the structured token space:
\begin{equation}
\hat{Z}_{t+h}=W_{\mathrm{out}}Y^L_h.
\end{equation}
The prediction heads decode $\hat{Z}_{t+h}$ into auxiliary future state predictions and the future policy target:
\begin{equation}
\hat{x}^{\mathrm{link}}_{t+h},
\hat{u}^{\mathrm{link}}_{t+h},
\hat{v}^{\mathrm{link}}_{t+h},
\hat{x}^{\mathrm{scene}}_{t+h},
\hat{u}^{\mathrm{scene}}_{t+h},
\hat{v}^{\mathrm{scene}}_{t+h},
\hat{g}_{t+h}
=
D_{\theta}(\hat{Z}_{t+h}).
\end{equation}
In our implementation, the transformer hidden dimension is $384$, the structured token dimension is $256$, and the default horizon set contains $9$ prediction horizons. The future query slots are initialized from the current structured token set and then transformed by horizon-conditioned blocks.

\subsubsection{Target Representation}
\label{supp:target_representation}

The decoded target $\hat{g}_{t+h}$ is designed to match the semantic interface consumed by the target-conditioned dexterous policy. It is a bimanual target,
\begin{equation}
\hat{g}_{t+h}=[\hat{g}^{R}_{t+h},\hat{g}^{L}_{t+h}],
\end{equation}
where each side contains future wrist information, hand-link or joint information, object motion information, fingertip relation terms, and shape-level task cues. In our implementation, to follow a fair setup to ManipTrans~\cite{maniptrans} and PhysGraph~\cite{physgraph}, the full target is $900$-dimensional, consisting of two $450$-dimensional hand-side targets. This includes wrist pose and velocity, joint delta positions and velocities, object pose and velocity, fingertip distance terms, and BPS shape features.

This target is not meant to be a physically exact rollout. It is a structured future coarse guidance for the policy. The downstream policy still observes the current state at every control step and performs contact-level correction through feedback control.

\subsubsection{Predictor Training Objective}
\label{supp:predictor_loss}

The predictor is trained by supervised future prediction from demonstration replay. For each sampled time $t$, the model predicts $\hat{Z}_{t+h}$ and $\hat{g}_{t+h}$ for all $h\in\mathcal{H}$. The loss is
\begin{equation}
\mathcal{L}_{\mathrm{pred}}
=
\lambda_z\mathcal{L}_z
+
\lambda_{\mathrm{state}}\mathcal{L}_{\mathrm{state}}
+
\lambda_{\mathrm{target}}\mathcal{L}_{\mathrm{target}}.
\end{equation}
The latent consistency loss stabilizes horizon-latent prediction:
\begin{equation}
\mathcal{L}_z
=
\|\hat{Z}_{t}-Z_t\|.
\end{equation}
The structured-state loss supervises hand-link and scene predictions:
\begin{equation}
\mathcal{L}_{\mathrm{state}}
=
\sum_{h\in\mathcal{H}}
\left[
\lambda_x\|\hat{x}_{t+h}-x_{t+h}\|
+
\lambda_u\|\hat{u}_{t+h}-u_{t+h}\|
+
\lambda_v\|\hat{v}_{t+h}-v_{t+h}\|
\right],
\end{equation}
where the terms are applied to both hand-link and scene-level predictions with separate weights.

The target loss is a component-wise loss over the target representation:
\begin{equation}
\mathcal{L}_{\mathrm{target}}
=
\sum_{h\in\mathcal{H}}
\sum_{b\in\mathcal{B}}
\lambda_b
d_b
\left(
\hat{g}^{b}_{t+h},
g^{\mathrm{demo},b}_{t+h}
\right),
\end{equation}
where $\mathcal{B}$ indexes target components such as wrist, link, object, fingertip, and shape terms. Most components use Smooth-$L_1$ distance. For quaternion components, we normalize both predicted and ground-truth quaternions and align their sign hemisphere before computing the loss, since $q$ and $-q$ represent the same rotation.


\subsubsection{Target-Conditioned Structured Dexterous Policy}
\label{supp:policy}

The dexterous policy receives the current simulator state $s_t$ and a target $g_t$. The target can either be the privileged demonstration target $g^{\mathrm{demo}}_t$ or the DexFuture-predicted target $\hat{g}_t$. Following PhysGraph \cite{physgraph}, the policy groups $s_t$ and $\hat{g}_t$ to structured hand-link inputs, then tokenizes the bimanual system hand-link inputs into hand-link tokens, scene tokens, and a policy token:
\begin{equation}
H^0_t
=
\{h^{\mathrm{link}}_{t,i}\}_{i=1}^{N_p}
\cup
\{h^{\mathrm{tool}}_t,h^{\mathrm{obj}}_t,h^{\mathrm{pol}}_t\}.
\end{equation}
A transformer encoder produces final tokens $H^L_t$. The policy token $h^{L,\mathrm{pol}}_t$ parameterizes a Gaussian action distribution:
\begin{equation}
a_t\sim
\mathcal{N}
\left(
\mu_\phi(h^{L,\mathrm{pol}}_t),
\mathrm{diag}(\sigma_\phi^2)
\right).
\end{equation}
The value function uses the policy token and training-time privileged features:
\begin{equation}
V_\phi(s_t)=f_V(h^{L,\mathrm{pol}}_t,s^{\mathrm{priv}}_t).
\end{equation}

Inspired by Physgraph \cite{physgraph}, we describe the controller as a target-conditioned per-link transformer policy, which can be replaced by any target-referenced based policy. The method contribution of this paper is the future-target predictor and the hierarchical target-generation pipeline, rather than the design of low-level controller.

\subsubsection{PPO Reward}
\label{supp:ppo_reward}

The policy is trained with PPO using imitation-style rewards. Let $s^{\mathrm{demo}}_t$ be the demonstration state aligned to the current progress index. The reward is a weighted sum of exponential tracking terms and regularization:
\begin{equation}
r_t
=
\sum_{m\in\mathcal{M}}
\beta_m
\exp
\left(
-\alpha_m d_m(s_t,s^{\mathrm{demo}}_t)
\right)
-
\beta_E\|a_t\|^2.
\end{equation}
The set $\mathcal{M}$ includes wrist position and rotation, fingertip or link position, object position and rotation, object linear and angular velocity, wrist velocity, and joint velocity terms. For bimanual tasks, rewards from the two hands are summed, and success requires both sides to satisfy the task-specific success criterion.

We conducted two phases of PPO training. Since the visuomotor predictor is always frozen, we always leverage the predicted targets rather than the demonstration for training. However, the input RGB to the visuomotor predictor are separate to two stages. In the stage one, we only use the offline causal RGB from demonstration to stabilize the policy training. In the second stage, we switch the RGB input to the online causal rendered RGB from rollout rather than offline demonstration, this enables the full hierarchical system to be fully closed-loop.

\subsubsection{Receding-Horizon Target Execution}
\label{supp:receding_horizon}

During execution, the predictor runs at a slower semantic timescale than the policy. At refresh time $t_j$, it predicts a sparse target sequence
\begin{equation}
\hat{\mathbf{g}}_{t_j:t_j+H}
=
F_\theta(\mathcal{O}_{t_j-K:t_j};\mathcal{H}).
\end{equation}
For an intermediate control step $t_j+\delta$, the target is obtained by linear interpolation. Let $h_a\leq\delta\leq h_b$ be neighboring horizons in $\mathcal{H}$. Then
\begin{equation}
\tilde{g}_{t_j+\delta}
=
(1-\eta)\hat{g}_{t_j+h_a}
+
\eta \hat{g}_{t_j+h_b},
\qquad
\eta=\frac{\delta-h_a}{h_b-h_a}.
\end{equation}
The policy then acts as
\begin{equation}
a_{t_j+\delta}
\sim
\pi_\phi(\cdot\mid s_{t_j+\delta},\tilde{g}_{t_j+\delta}).
\end{equation}
This allows the target predictor to produce coarse future-state guidance over a window, while the policy executes high-frequency feedback control at every simulator step.

In the default setting, the history length is $K=8$ and the horizon set is
\begin{equation}
\mathcal{H}=\{0,2,4,6,8,10,12,14,16\}.
\end{equation}
Thus, at refresh time $t_j$, the predictor consumes observations from
\[
\mathcal{O}_{t_j-8:t_j}
=
\{I_{\tau},p_{\tau}\}_{\tau=t_j-8}^{t_j},
\]
and predicts sparse future targets
\[
\{\hat{g}_{t_j},\hat{g}_{t_j+2},\hat{g}_{t_j+4},\ldots,\hat{g}_{t_j+16}\}.
\]
The policy acts at every simulator step, so targets for intermediate steps such as $t_j+1,t_j+3,\ldots,t_j+15$ are obtained by linear interpolation between neighboring sparse predictions.

\subsection{Pseudocode}
\begin{algorithm}[t]
\caption{Training the Future-State Visuomotor Target Predictor}
\label{alg:train_predictor}
\begin{algorithmic}[1]
\Require Demonstration dataset $\mathcal{D}$, frozen visual encoder $\Psi$, horizon set $\mathcal{H}$, history length $K$, predictor $F_\theta$, loss weights $\lambda$.
\Ensure Trained predictor $F_\theta$.

\While{not converged}
    \State Sample a minibatch of time indices $t$ and windows from $\mathcal{D}$:
    \[
    \{I_{\tau},p_{\tau}\}_{\tau=t-K}^{t},
    \qquad
    \{g^{\mathrm{demo}}_{t+h}\}_{h\in\mathcal{H}}.
    \]

    \For{$\tau=t-K,\ldots,t$}
        \State Encode RGB observation:
        \[
        V^{\mathrm{raw}}_{\tau} \gets \Psi(I_{\tau}).
        \]
        \State Project visual tokens into the structured token space:
        \[
        V_{\tau} \gets W_vV^{\mathrm{raw}}_{\tau}+E_{\mathrm{patch}}.
        \]
        \State Construct structured visuomotor tokens:
        \[
        Z_{\tau}\gets \mathrm{Tokenize}(V_{\tau},p_{\tau}),
        \]
        where $\mathrm{Tokenize}(\cdot)$ builds hand-link tokens by local visual cross-attention and tool/object tokens by anchor-aligned scene pooling.
    \EndFor

    \State Form the observed structured history:
    \[
    Z_{t-K:t}\gets \{Z_{t-K},\ldots,Z_t\}.
    \]

    \For{$h\in\mathcal{H}$}
        \State Initialize future query tokens:
        \[
        Y^0_h \gets W_{\mathrm{in}}Z_t+E_{\mathrm{slot}}+E_{\mathrm{frame}}(h).
        \]
        \State Compute horizon condition:
        \[
        c_h\gets f_h(\gamma(h)).
        \]
    \EndFor

    \For{$\ell=1,\ldots,L$}
        \For{$h\in\mathcal{H}$}
            \State Update future query tokens:
            \[
            Y^\ell_h
            \gets
            \mathrm{HCTBlock}
            \left(
            Y^{\ell-1}_h,
            W_{\mathrm{in}}Z_{t-K:t},
            c_h
            \right).
            \]
        \EndFor
    \EndFor

    \For{$h\in\mathcal{H}$}
        \State Decode future structured tokens:
        \[
        \hat{Z}_{t+h}\gets W_{\mathrm{out}}Y^L_h.
        \]
        \State Decode future policy target:
        \[
        \hat{g}_{t+h}\gets D_\theta(\hat{Z}_{t+h}).
        \]
    \EndFor

    \State Compute prediction objective:
    \[
    \mathcal{L}_{\mathrm{pred}}
    \gets
    \lambda_z\mathcal{L}_z
    +
    \lambda_{\mathrm{state}}\mathcal{L}_{\mathrm{state}}
    +
    \lambda_{\mathrm{target}}\mathcal{L}_{\mathrm{target}}.
    \]

    \State Update predictor parameters:
    \[
    \theta\gets\theta-\eta\nabla_\theta\mathcal{L}_{\mathrm{pred}}.
    \]
\EndWhile
\end{algorithmic}
\end{algorithm}

\begin{algorithm}[t]
\caption{Training the Target-Conditioned Structured Dexterous Policy}
\label{alg:train_policy}
\begin{algorithmic}[1]
\Require Simulator environment, demonstration dataset $\mathcal{D}$, target source $\mathcal{G}$, policy $\pi_\phi$, PPO optimizer.
\Ensure Trained target-conditioned policy $\pi_\phi$.

\While{not converged}
    \State Reset parallel environments to demonstration-aligned initial states.

    \For{rollout step $t=1,\ldots,T_{\mathrm{roll}}$}
        \State Read current simulator state $s_t$.

        \If{privileged target mode}
            \State Obtain demonstration target:
            \[
            \tilde{g}_t \gets g^{\mathrm{demo}}_{t+h}.
            \]
        \Else
            \State Obtain predicted target from target cache:
            \[
            \tilde{g}_t
            \gets
            \mathrm{Interp}
            \left(
            \hat{\mathbf{g}}_{t_j:t_j+H},
            t-t_j
            \right).
            \]
        \EndIf

        \State Tokenize current state and target:
        \[
        H^0_t\gets \mathrm{PolicyTokenize}(s_t,\tilde{g}_t).
        \]

        \State Compute policy distribution:
        \[
        \pi_\phi(\cdot\mid s_t,\tilde{g}_t)
        \gets
        \mathrm{PolicyTransformer}(H^0_t).
        \]

        \State Sample action:
        \[
        a_t\sim \pi_\phi(\cdot\mid s_t,\tilde{g}_t).
        \]

        \State Step simulator:
        \[
        s_{t+1}\gets \mathrm{EnvStep}(s_t,a_t).
        \]

        \State Compute imitation-style reward:
        \[
        r_t
        \gets
        \sum_{m\in\mathcal{M}}
        \beta_m
        \exp
        \left(
        -\alpha_m d_m(s_t,s^{\mathrm{demo}}_t)
        \right)
        -
        \beta_E\|a_t\|^2.
        \]

        \State Store transition:
        \[
        (s_t,\tilde{g}_t,a_t,r_t,s_{t+1}).
        \]
    \EndFor

    \State Update $\phi$ with PPO using collected rollouts.
\EndWhile
\end{algorithmic}
\end{algorithm}

\begin{algorithm}[t]
\caption{DexFuture Receding-Horizon Execution}
\label{alg:execution}
\begin{algorithmic}[1]
\Require Trained predictor $F_\theta$, trained policy $\pi_\phi$, horizon set $\mathcal{H}$, history length $K$, predictor refresh stride $S$.
\Ensure Executed bimanual manipulation trajectory.

\State Initialize observation history buffer $\mathcal{O}_{t-K:t}$.

\For{refresh time $t_j=0,S,2S,\ldots$}
    \State Predict sparse future target sequence:
    \[
    \hat{\mathbf{g}}_{t_j:t_j+H}
    \gets
    F_\theta(\mathcal{O}_{t_j-K:t_j};\mathcal{H}).
    \]

    \For{$\delta=0,\ldots,S-1$}
        \State Interpolate target for the current control step:
        \[
        \tilde{g}_{t_j+\delta}
        \gets
        \mathrm{Interp}
        \left(
        \hat{\mathbf{g}}_{t_j:t_j+H},
        \delta
        \right).
        \]

        \State Read current simulator state $s_{t_j+\delta}$.

        \State Query target-conditioned policy:
        \[
        a_{t_j+\delta}
        \sim
        \pi_\phi
        \left(
        \cdot
        \mid
        s_{t_j+\delta},
        \tilde{g}_{t_j+\delta}
        \right).
        \]

        \State Step simulator:
        \[
        s_{t_j+\delta+1}
        \gets
        \mathrm{EnvStep}
        \left(
        s_{t_j+\delta},
        a_{t_j+\delta}
        \right).
        \]

        \State Record the new RGB/state observation and update the history buffer $\mathcal{O}$.
    \EndFor
\EndFor
\end{algorithmic}
\end{algorithm}

\end{document}